\documentclass[10pt,twocolumn,letterpaper]{article}

\usepackage{dicta}
\usepackage{times}
\usepackage{epsfig}
\usepackage{amsmath}
\usepackage{amssymb}
\usepackage{booktabs}
\usepackage{graphicx}
\usepackage{multirow}
\usepackage{xcolor}
\usepackage{placeins}
\usepackage{float}
\usepackage{xfp}
\usepackage{etoolbox}
\usepackage{lastpage}
\usepackage{refcount}
\usepackage{tikz}
\usetikzlibrary{arrows.meta,calc,positioning}
\usepackage[pagebackref=false,breaklinks=true,letterpaper=true,colorlinks,bookmarks=false]{hyperref}

\newif\ifshowauthors
\showauthorsfalse
\ifshowauthors
\dictafinalcopy
\fi

\dictafinalcopy

\def\dictaPaperID{67}

\newcommand{\metricformat}[3]{%
  \ifnum\fpeval{abs((#1)-(#3)) < 0.00005}=1
    \ifnum\fpeval{#2}=1 \textbf{\underline{#1}}\else \underline{#1}\fi
  \else
    \ifnum\fpeval{#2}=1 \textbf{#1}\else #1\fi
  \fi}
\newcommand{\metriccd}[3]{\metricformat{#1}{(#1)<(#2)}{#3}}
\newcommand{\metricf}[3]{\metricformat{#1}{(#1)>(#2)}{#3}}
\newcommand{\metricbest}[2]{%
  \ifnum\fpeval{abs((#1)-(#2)) < 0.00005}=1 \underline{#1}\else #1\fi}
\newcommand{\resulttablesize}{\scriptsize}
\newcommand{\resulttablespacing}[1]{\setlength{\tabcolsep}{#1}\renewcommand{\arraystretch}{0.92}}

\ifshowauthors\fi

\title{Target-Part Preference Optimization for Part-Aware 3D Generation}
\showauthorstrue
\ifshowauthors
\author{
\begin{tabular}{c}
{\large Zishan Qin$^{1,2}$ \quad Rui Li$^{3}$ \quad Zeyu Zhang$^{1}$ \quad Sevvandi Kandanaarachchi$^{2}$ \quad Lina Yao$^{1}$ \quad Dong Gong$^{1,*}$}\\
$^{1}$UNSW Sydney\quad
$^{2}$CSIRO\quad
$^{3}$NTU \quad
 {\small $^*$Corresponding author}\\[2pt]
{\tt\small \
zishan.qin@unsw.edu.au \quad edgong01@gmail.com}
\end{tabular}
}
\else
\author{Anonymous DICTA 2026 Submission}
\fi

\begin{document}
\maketitle
\thispagestyle{empty}
\pagestyle{empty}


\begin{figure*}[t]
\centering
\resizebox{0.95\textwidth}{!}{\input{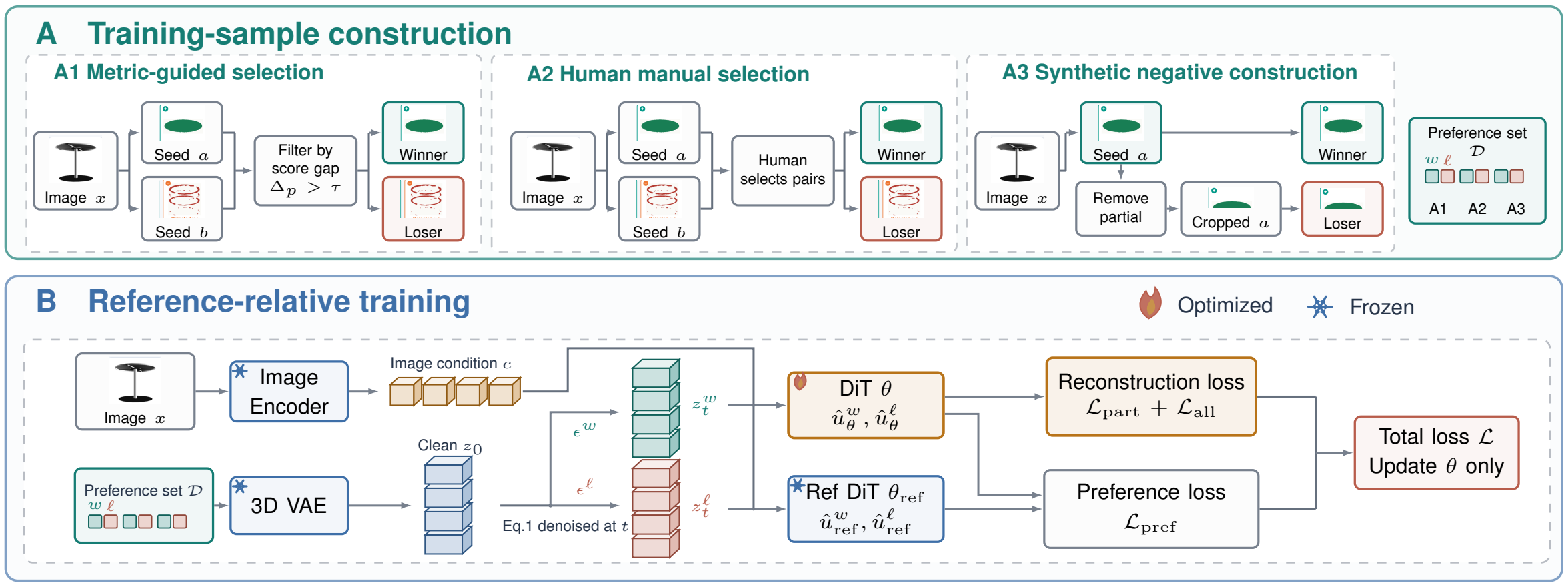}}
\caption{Overview of target-part preference fine-tuning. Superscripts $w$ and $\ell$ denote winner and loser trajectories. (A) For target part $p$,
GT-referenced metrics rank candidates and order their seeds, which instantiate different noise
trajectories toward the same GT clean latent. Human controls rank candidates visually. Synthetic-crop controls remove loser target geometry before variational autoencoder (VAE) encoding.
All three constructions write ordered $(w,\ell)$ pairs into one preference set $\mathcal D$. (B) $\mathcal D$ is the training input. In the primary construction, the 3D VAE encodes one clean latent $z_0$, which is noised at a shared timestep $t$ with $\epsilon^w$ and $\epsilon^{\ell}$ (Eq.~\ref{eq:flow_path}). Only the PartCrafter Diffusion Transformer (DiT) $\theta$ is optimized with the total loss $\mathcal L$ in Sec.~\ref{sec:preference_objective}. The image encoder, 3D VAE, and stage-initial reference DiT remain frozen.}
\label{fig:pipeline_overview}
\end{figure*}

\begin{abstract}
Part-aware 3D generation requires not only globally plausible objects but also complete, well-separated semantic parts, yet high-quality part decompositions remain too scarce to support large-scale supervised fine-tuning. We study whether preference fine-tuning based on Direct Preference Optimization (DPO) can adapt a pretrained part-aware generator from limited part-annotated data without architectural changes.
We rank candidates generated with different random seeds against ground-truth (GT) geometric metrics, and select the (object, target-part) examples with the largest across-seed quality gap. Each selected example stores a seed pair whose two flow trajectories share one clean GT compositional latent and differ only in Gaussian noise.
We derive a reference-relative target-block preference loss from FlowDPO that compares the two trajectories against a frozen reference model. GT flow-matching losses on the target block and complete part set provide absolute supervision, while only the diffusion transformer (DiT) is updated. Human-ranked controls test an alternative ordering source, and synthetic cropping provides controlled loser targets that are incomplete by construction. Among the three training-set scales, the medium-scale setting gives the strongest result on held-out Objaverse and outperforms both the pretrained generator and the reported direct supervised fine-tuning baseline on the multi-part aggregate. Further analysis shows sensitivity to pair ordering, target-part coverage, and the balance between pairwise and GT supervision. Cross-dataset experiments show less consistent gains under distribution shift.
\end{abstract}


\section{Introduction}

Image- and text-conditioned 3D generators can now synthesize objects with convincing global geometry \cite{poole2022dreamfusion,liu2023zero1to3,hong2023lrm}. Many applications, however, manipulate objects at the part level. Editing a component, recomposing an object, or articulating a joint requires the relevant semantic parts to be complete, well separated, and independently usable \cite{chen2025partgen,lin2025partcrafter,liu2024singapo,liu2026pact}. A plausible overall shape does not guarantee these properties. Generated parts may still be missing, fused, or fragmented. Such local failures may be obscured by object-level evaluation but remain important for downstream use, motivating supervision that explicitly targets part quality.

Supervised fine-tuning (SFT) on complete part decompositions is a direct solution, but fine-grained 3D part annotations are costly and remain limited \cite{mo2019partnet,yang2024sampart3d}. Making effective use of a limited number of annotated objects is therefore important. Preference optimization provides a complementary source of supervision. Direct Preference Optimization (DPO) learns from ordered comparisons relative to a frozen reference model without training a separate reward model \cite{rafailov2023direct}. Diffusion-DPO and FlowDPO extend this formulation to diffusion and flow-matching generators by replacing likelihood ratios with denoising- or flow-matching-error surrogates \cite{wallace2024diffusiondpo,jiao2024flowdpo}. We ask whether preference-guided fine-tuning can adapt a pretrained part-aware generator from limited annotated data without degrading its object-level generation ability.

Applying DPO to part-aware 3D generation is not straightforward. Existing 3D preference methods rank complete assets or rendered views \cite{zhou2025dreamdpoaligningtextto3dgeneration,li2026circulardpo}, while Mesh-RFT localizes feedback through quality-aware face masking \cite{liu2025meshrftenhancingmeshgeneration}. These formulations do not directly address comparisons on one semantic part block in a generator that jointly models several parts. Existing part datasets likewise provide decompositions, not ordered target-part preferences \cite{mo2019partnet,yang2024sampart3d}. This leaves two practical questions: how to construct reliable target-part pairs from limited annotations, and how to improve the selected part without degrading the overall object quality.

We customize FlowDPO for target-part preference learning and fine-tune only PartCrafter's diffusion transformer (DiT) without architectural changes \cite{lin2025partcrafter,jiao2024flowdpo}. PartCrafter represents each part with a separate latent-token block, allowing the pairwise error to be evaluated on a selected target block. For each target part,
we rank candidates sampled from the frozen generator using GT-referenced geometric metrics and keep the examples with the largest across-seed quality gap; each kept example stores a seed pair.
These seeds instantiate two noisy trajectories toward the same GT clean compositional latent. A reference-relative objective compares their target-block flow errors, while target-part and all-part GT losses provide absolute supervision. Among the tested scales, the medium-scale setting gives the strongest held-out Objaverse multi-part result, raising matched-part F-score by approximately 5\% over pretrained PartCrafter and exceeding the reported direct-SFT baseline.
We make three contributions.
\begin{itemize}
    \item We devise a GT-referenced ranking procedure that converts generated candidates into winner--loser seed pairs with varying seed-conditioned trajectories. We derive a target-block preference objective from FlowDPO \cite{jiao2024flowdpo} and combine it with target-part and all-part GT flow-matching losses.
    \item We construct metric-ranked target-part training sets at three scales from PartObjaverse-Tiny \cite{yang2024sampart3d}, together with human-ranked and synthetic-crop controls.
    \item We analyze training-set scale, target-part coverage, ordering reliability, loser construction, objective components, and cross-dataset transfer to identify when target-part preference learning helps and where its gains do not transfer.
\end{itemize}

\section{Related Work}
\subsection{3D And Part-Aware Generation}
Modern 3D generation includes score-distillation methods guided by pretrained image models \cite{poole2022dreamfusion,lin2023magic3d}, sparse-view reconstruction \cite{liu2023zero1to3}, and feed-forward generators that predict 3D representations directly from images \cite{hong2023lrm,li2025triposg}. These approaches substantially improve object-level fidelity, but a plausible object surface does not by itself provide explicit, separately usable components.

To generate better part components, part-aware methods generate or recover component-level structure through combining segmentation and reconstruction \cite{chen2025partgen}, compositional or contextual part latents \cite{lin2025partcrafter,dong2025from}, autoregressive part discovery \cite{chen2025autopartgen}, semantic decoupling and structural cohesion \cite{yang2025omnipart}, efficient multi-part packing \cite{tang2025efficient}, or retrieval-augmented generation and editing \cite{li2026partrag}. Explicit parts also support downstream articulation, where methods infer part graphs, joints, axes, or motion ranges \cite{liu2024singapo,liu2026pact}. Our contribution is an alignment objective rather than a new 3D representation. We evaluate this objective on PartCrafter \cite{lin2025partcrafter}.

\subsection{Preference Optimization For 3D Generation}
Direct Preference Optimization (DPO) aligns a model from preference pairs relative to a fixed reference without training a separate reward model \cite{rafailov2023direct}. Diffusion-DPO and FlowDPO extend this formulation to diffusion and flow models by replacing intractable output likelihoods with denoising- or flow-matching-error surrogates \cite{wallace2024diffusiondpo,jiao2024flowdpo}. Because such objectives depend on reliable preference labels, robust-DPO work also studies learning under noisy feedback \cite{raychowdhury2024robustdpo}.

Preference alignment has recently been applied to 3D generation. DreamDPO ranks paired renderings with reward-model or multimodal-model feedback during text-to-3D optimization \cite{zhou2025dreamdpoaligningtextto3dgeneration}. Circular-DPO transfers preference information across stages of a 3D generation pipeline \cite{li2026circulardpo}, while Mesh-RFT applies fine-grained reinforcement feedback to mesh generation \cite{liu2025meshrftenhancingmeshgeneration}. DreamDPO and Circular-DPO primarily rank complete objects or rendered views, whereas Mesh-RFT provides fine-grained mesh feedback. Our study instead compares reference-relative flow errors on one target-part latent block. Section~\ref{sec:problem_setup} defines the metric-ranked construction and output-ranked controls.

\section{Preliminaries}
\label{sec:preliminaries}
We first summarize the PartCrafter backbone and the flow-matching notation used in this paper.

\subsection{PartCrafter Backbone}
\label{sec:partcrafter_backbone}
PartCrafter jointly generates a requested number $P$ of part meshes from one RGB image \cite{lin2025partcrafter}. It inherits the TripoSG image encoder, variational autoencoder (VAE), and Diffusion Transformer (DiT) \cite{li2025triposg}. For input image $x$, the image encoder produces visual tokens $h_x=E_{\mathrm{img}}(x)$, denoted by $c$ in our loss.

Let $\mathcal G=\{g_q\}_{q=1}^{P}$ be the ground-truth part surfaces. The VAE represents each surface $g_q$ by a block of $N$ latent tokens with feature width $d$. We denote this block by $z_{0,q}\in\mathbb R^{N\times d}$ and the complete compositional latent by $z_0=(z_{0,1},\ldots,z_{0,P})\in\mathbb R^{P\times N\times d}$. Learnable part identity embeddings distinguish the $P$ token groups. The DiT alternates local attention within each part with global attention across all part tokens, and the shared VAE decoder maps each latent block to a part mesh. We use $q$ for a generic part and reserve $p$ for the part ranked and evaluated in one training example. Accordingly, $z_{0,p}$ denotes the latent token block associated with target part $p$. The index $p$ guides offline ranking and selects the corresponding block in the loss. It is not a semantic class label or generator condition.

\subsection{Flow Matching And FlowDPO}
\label{sec:flowdpo_preliminaries}
Following PartCrafter's rectified flow \cite{lin2025partcrafter,liu2023rectifiedflow}, let $i\in\{w,\ell\}$ index the winner and loser trajectories of one stored preference pair. For either trajectory, $z_0^i,\epsilon^i,z_t^i,u^i\in\mathbb R^{P\times N\times d}$ denote the clean compositional latent, Gaussian noise, noised latent, and reverse-velocity target, respectively. The scheduler form is
\begin{equation}
z_t^i=(1-\sigma_t)z_0^i+\sigma_t\epsilon^i,
\qquad u^i=z_0^i-\epsilon^i,
\label{eq:flow_path}
\end{equation}
where $\epsilon^i\sim\mathcal N(0,I)$ has the same shape as $z_0^i$. In Circular-DPO notation \cite{li2026circulardpo}, Eq.~\ref{eq:flow_path} follows the same straight path with endpoints $x_0^i=\epsilon^i$, $x_1^i=z_0^i$, and interpolation coordinate $\tau=1-\sigma_t$. Only the scheduler parameterization differs. Let $\xi^i=(z_0^i,\epsilon^i,c)$ collect one trajectory's clean latent, noise, and image conditioning. The mean-squared flow-matching error (MSE) over the $P\times N\times d$ compositional latent is
\begin{equation}
\operatorname{MSE}_t(\xi^i;\theta)
=\frac{1}{P\,N\,d}\left\|\hat u_{\theta}(z_t^i,t;c)-u^i\right\|_F^2.
\label{eq:generic_error}
\end{equation}
Standard conditional flow matching minimizes $\mathcal L_{\mathrm{CFM}}(\theta)=\mathbb E_{\xi,t}[\operatorname{MSE}_t(\xi;\theta)]$ over clean latents, noise, and timesteps \cite{lipman2023flow}.

FlowDPO and Circular-DPO replace the output log-likelihood ratio with a flow-matching-error surrogate and apply a log-sigmoid to the winner-minus-loser difference of reference-relative MSE reductions \cite{jiao2024flowdpo,li2026circulardpo}. We retain this $\beta$-scaled structure and define the target-block error in Eq.~\ref{eq:target_error}. FlowDPO and Circular-DPO normally compare ordered clean samples. In our primary metric-ranked construction, both trajectories instead share one GT clean latent and differ only in seed-determined noise; offline candidate ranking fixes their winner/loser order. The frozen $\theta_{\mathrm{ref}}$ is the DiT at the start of the current stage; both trajectories share $c$ and $t$.

\section{Method}

Our goal is to adapt the pretrained part generator from a limited set of GT part decompositions so that it improves a selected part while preserving the remaining parts. PartCrafter represents each part as a separate latent-token block under a shared DiT, allowing the FlowDPO error surrogate to be evaluated on target block $p$. This localizes the training signal, although the parameter update still affects the shared DiT. Three ingredients realize it: (1) preference pairs built offline by ranking seed-generated candidates with GT part metrics, so the winner and loser trajectories share one GT clean latent and differ only in seeded noise; (2) a reference-relative loss on block $p$ that favors the winner trajectory; and (3) target-part and all-part reconstruction on the winner that supply absolute GT supervision and keep the surrounding object faithful. Sections~\ref{sec:problem_setup}--\ref{sec:pair_construction} detail the pair construction and its controls, and Sec.~\ref{sec:preference_objective} presents the objective.

\subsection{Metric-Supervised Trajectory Pairs}
\label{sec:problem_setup}
A training example is $(x,\mathcal G,p,s^w,s^{\ell})$, containing the input image, GT part set, target-part index, and stored winner and loser random seeds. Before fine-tuning, the pair builder samples multiple candidates for the same image and requested part count. It evaluates each candidate's indexed part $p$ against $g_p$ with the automatic score in supplementary Sec.~S2. For each retained pair, the higher- and lower-scoring candidates determine $s^w$ and $s^{\ell}$, which differ as seed-conditioned trajectories rather than as two encoded candidate meshes. The trainer consumes this stored order rather than recomputing candidate-quality scores. The GT part annotation supplies the clean geometry used by the flow-matching targets.

Encoding $\mathcal G$ gives compositional latent $z_0\in\mathbb R^{P\times N\times d}$. The stored seeds deterministically instantiate noise tensors $\epsilon^w$ and $\epsilon^{\ell}$ of the same shape, producing two noised latent states from the same GT compositional latent.
 The primary construction therefore compares seed-conditioned stochastic trajectories toward one target geometry rather than two encoded candidate meshes. Candidate ranking is used only offline to decide the winner/loser order; GT part geometry remains the clean training target. In the human-ranked controls, visual inspection labels the seeds associated with two generated candidates. Synthetic-crop controls instead crop the loser target surface and independently re-encode the complete loser part set.
 The non-target surfaces are unchanged, but their stochastic VAE latent samples need not be numerically identical across the two trajectories.

\subsection{Pair Construction And Controls}
\label{sec:pair_construction}
The primary construction yields one winner/loser seed pair for each selected target part. Controls introduce visual ordering through human ranking, change the loser clean geometry through synthetic cropping, or perturb stored pairs through label reversal and random-pair augmentation. Section~\ref{sec:experiment_protocol} and supplementary Sec.~S2 give the dataset scales, intervention ratios, and configurations.

\subsection{Target-Block Preference Objective}
\label{sec:preference_objective}
For each ordered pair, Eq.~\ref{eq:flow_path} is instantiated separately for $i\in\{w,\ell\}$. The generic error in Eq.~\ref{eq:generic_error} averages the squared error over all $P$ part blocks, with $N$ tokens and $d$ features in each block. We restrict the preference comparison to the tokens and features of target block $p$.
\begin{equation}
\operatorname{MSE}_{t}^{p}(\xi^i;\theta)
=\frac{1}{N\,d}\left\|\hat u_{\theta,p}(z_t^i,t;c)-u_p^i\right\|_F^2.
\label{eq:target_error}
\end{equation}
For any part $q\in\{1,\ldots,P\}$, $\operatorname{MSE}_{t}^{q}$ is defined by replacing $p$ with $q$ in Eq.~\ref{eq:target_error}.
For each trajectory, the trainable and reference models receive the same $(z_t^i,c,t)$. Let $r_i$ denote the reference-relative reduction in target-part error. A positive $r_i$ means that the trainable DiT predicts trajectory $i$ more accurately than the frozen reference at that timestep. In Eq.~\ref{eq:pair}, $\sigma(\cdot)$ without a timestep subscript is the logistic sigmoid and is distinct from the scheduler level $\sigma_t$ in Eq.~\ref{eq:flow_path}. As in Circular-DPO, the per-example, per-timestep pair loss is
{\small
\begin{equation}
\begin{aligned}
r_i &= \operatorname{MSE}_{t}^{p}(\xi^i;\theta_{\mathrm{ref}})-\operatorname{MSE}_{t}^{p}(\xi^i;\theta), \qquad i\in\{w,\ell\},\\
\ell_{\mathrm{pair}} &= -\log\sigma\!\left(\beta\left[r_w-r_\ell\right]\right).
\end{aligned}
\label{eq:pair}
\end{equation}
}
Equation~\ref{eq:pair} keeps Circular-DPO's reference-relative winner-minus-loser structure but replaces its full-latent flow-matching error with target-block error. Minimizing it increases $r_w-r_\ell$. Relative to the frozen reference, the trainable DiT is encouraged to reduce target-part error more on the winner trajectory than on the loser trajectory. It does not independently require $\operatorname{MSE}_{t}^{p}(\xi^\ell;\theta)<\operatorname{MSE}_{t}^{p}(\xi^\ell;\theta_{\mathrm{ref}})$. Its gradient is $\nabla_\theta\ell_{\mathrm{pair}}=\beta\sigma\!\left[-\beta(r_w-r_\ell)\right](\nabla_\theta\operatorname{MSE}_{t}^{p}(\xi^w;\theta)-\nabla_\theta\operatorname{MSE}_{t}^{p}(\xi^\ell;\theta))$, which is generally nonzero even when the primary pair shares $z_0$ because its two noises produce different inputs and velocity targets.

Two reconstruction terms on the winner trajectory retain absolute GT supervision. Let $\mathcal L_{\mathrm{pair}}=\mathbb E[\ell_{\mathrm{pair}}]$ and $\mathcal L_{\mathrm{target}}=\mathbb E[\operatorname{MSE}_{t}^{p}(\xi^w;\theta)]$ denote expectations over training examples and sampled timesteps. We compute $\mathcal L_{\mathrm{all}}$ by averaging $\operatorname{MSE}_{t}^{q}(\xi^w;\theta)$ over every part block $q$ of every winner example. An example with $P$ parts therefore contributes $P$ blocks. In the reported primary configurations, each stored pair has unit example weight and pair margins are disabled. The implemented optimization schedule is
{\small
\begin{equation}
\mathcal L_s(\theta)=
\begin{cases}
\mathcal L_{\mathrm{target}}+\lambda_{\mathrm{all}}\mathcal L_{\mathrm{all}}, & s<S_{\mathrm{warm}},\\
\lambda_{\mathrm{pair}}\mathcal L_{\mathrm{pair}}+\lambda_{\mathrm{part}}\mathcal L_{\mathrm{target}}+\lambda_{\mathrm{all}}\mathcal L_{\mathrm{all}}, & s\ge S_{\mathrm{warm}}.
\end{cases}
\label{eq:training_schedule}
\end{equation}
}
Here $s$ without a superscript is the update index, distinct from the stored seeds $s^w$ and $s^\ell$, and $S_{\mathrm{warm}}$ is the reconstruction-only warm-up duration. Equation~\ref{eq:training_schedule} switches once from reconstruction-only warm-up to the combined objective. Its coefficients are fixed within each phase. Before the switch, the pairwise term is absent and target reconstruction has coefficient one. Afterwards, the winner trajectory enters the pairwise term and both reconstruction terms, whereas the loser trajectory enters only the pairwise term. The reconstruction terms compare the trainable DiT with GT flow targets. They do not constrain its outputs or parameters to remain close to $\theta_{\mathrm{ref}}$. The frozen reference appears only in $\mathcal L_{\mathrm{pair}}$. Direct SFT instead uses standard flow matching without preference pairs. Supplementary Sec.~S2 reports the configuration values.


%
%
%
%

\section{Experiments}

\begin{figure}[!t]
  \centering
  \includegraphics[width=0.5\textwidth]{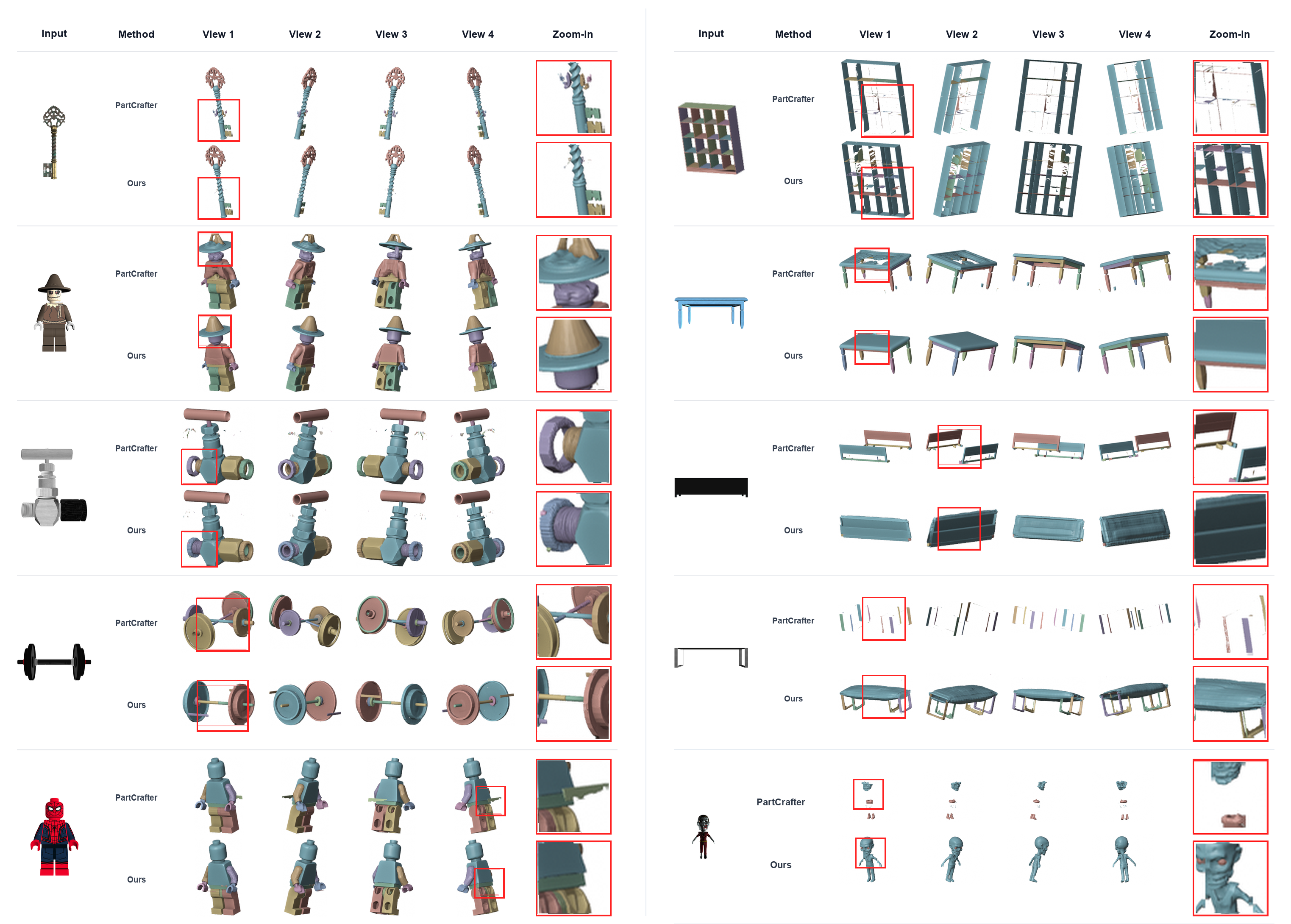}
  \caption{Qualitative effects of part-level alignment. Each comparison uses the same input image and requested part count for PartCrafter and Ours. Across articulated characters, furniture, and mechanical objects, Ours gives more complete target geometry and cleaner separation from neighboring parts; the zoomed regions highlight thin structures, contact boundaries, and small components.}
  \label{fig:prediction_examples_compact}
\end{figure}

\begin{table}[!t]
\centering
\caption{Comparison by input part count on held-out Objaverse~\cite{deitke2023objaverse}. We report object/matched-part Chamfer Distance (Obj-CD/Part-CD) and F-score (Obj-FS/Part-FS). The pretrained model is PartCrafter~\cite{lin2025partcrafter}. Ours is the large-scale setting; the medium setting appears as B2 in Table~\ref{tab:construction_controls}. SFT fine-tunes 140 GT objects with standard flow matching. The 2--8 average (avg.) covers valid multi-part examples. Supplementary Table~S2 gives complete cross-dataset results.}
\label{tab:baseline_controls}
\resulttablesize
\resulttablespacing{2.0pt}
\begin{tabular*}{\columnwidth}{@{\extracolsep{\fill}}llrrrr@{}}
\toprule
Part Count & Method & Obj-CD $\downarrow$ & Obj-FS $\uparrow$ & Part-CD $\downarrow$ & Part-FS $\uparrow$ \\
\midrule
\multirow{3}{*}{1 part} & Pretrained & \metricbest{0.2382}{0.2184} & \metricbest{0.6366}{0.6603} & \metricbest{0.2096}{0.1890} & \metricbest{0.6949}{0.7273} \\
 & SFT & \metriccd{0.2243}{0.2382}{0.2184} & \metricf{0.6441}{0.6366}{0.6603} & \metriccd{0.1890}{0.2096}{0.1890} & \metricf{0.7273}{0.6949}{0.7273} \\
 & \textbf{Ours} & \metriccd{0.2184}{0.2382}{0.2184} & \metricf{0.6603}{0.6366}{0.6603} & \metriccd{0.2044}{0.2096}{0.1890} & \metricf{0.7020}{0.6949}{0.7273} \\
\midrule
\multirow{3}{*}{2--4 parts} & Pretrained & \metricbest{0.2631}{0.2491} & \metricbest{0.5909}{0.6121} & \metricbest{0.2574}{0.2485} & \metricbest{0.5986}{0.6237} \\
 & SFT & \metriccd{0.2568}{0.2631}{0.2491} & \metricf{0.5945}{0.5909}{0.6121} & \metriccd{0.2620}{0.2574}{0.2485} & \metricf{0.5955}{0.5986}{0.6237} \\
 & \textbf{Ours} & \metriccd{0.2491}{0.2631}{0.2491} & \metricf{0.6121}{0.5909}{0.6121} & \metriccd{0.2485}{0.2574}{0.2485} & \metricf{0.6237}{0.5986}{0.6237} \\
\midrule
\multirow{3}{*}{5--8 parts} & Pretrained & \metricbest{0.2865}{0.2813} & \metricbest{0.5332}{0.5520} & \metricbest{0.3333}{0.3062} & \metricbest{0.4646}{0.4912} \\
 & SFT & \metriccd{0.2813}{0.2865}{0.2813} & \metricf{0.5463}{0.5332}{0.5520} & \metriccd{0.3197}{0.3333}{0.3062} & \metricf{0.4846}{0.4646}{0.4912} \\
 & \textbf{Ours} & \metriccd{0.2824}{0.2865}{0.2813} & \metricf{0.5520}{0.5332}{0.5520} & \metriccd{0.3062}{0.3333}{0.3062} & \metricf{0.4912}{0.4646}{0.4912} \\
\midrule
\multirow{3}{*}{\textbf{avg.}} & Pretrained & \metricbest{0.2724}{0.2625} & \metricbest{0.5680}{0.5881} & \metricbest{0.2875}{0.2716} & \metricbest{0.5453}{0.5707} \\
 & SFT & \metriccd{0.2667}{0.2724}{0.2625} & \metricf{0.5751}{0.5680}{0.5881} & \metriccd{0.2852}{0.2875}{0.2716} & \metricf{0.5509}{0.5453}{0.5707} \\
 & \textbf{Ours} & \metriccd{0.2625}{0.2724}{0.2625} & \metricf{0.5881}{0.5680}{0.5881} & \metriccd{0.2716}{0.2875}{0.2716} & \metricf{0.5707}{0.5453}{0.5707} \\
\bottomrule
\end{tabular*}
\end{table}

\subsection{Experimental Setup}
\label{sec:experiment_protocol}
\noindent\textbf{Training Data.}
We construct the main training set from PartObjaverse-Tiny \cite{yang2024sampart3d}. After retaining objects with two to eight usable parts and removing parts whose face fraction is below 0.005, the set contains 140 objects and 796 target-part pair examples. For each example,
the offline pair builder samples candidates and stores the winner/loser seeds, target-part index, and indexed GT geometry.
During fine-tuning, the seeds generate independent noise for the same GT compositional latent, as defined in Sec.~\ref{sec:problem_setup}. Candidate-quality metrics are used only during offline pair construction and are not optimized as rewards. The automatic score is a weighted sum of object FS/CD ($0.35/0.20$), target-part FS/CD
($0.20/0.10$), normal consistency ($0.10$), and component quality ($0.05$). Examples are ranked by the across-seed spread of
this score on the indexed part.
Supplementary Sec.~S2 gives the complete definition.

The training-set-size study spans three scales: \textcolor{black}{small} (56 pairs from 7 objects), \textcolor{black}{medium} (363 pairs from 50 objects), and \textcolor{black}{large} (796 pairs from 140 objects). The \textcolor{black}{small and medium} sets are selected from the ranked pool by usable part count, minimum part fraction, and category coverage. The \textcolor{black}{small and medium} checkpoints are first-stage runs, while the \textcolor{black}{large} set is a continuation checkpoint with refreshed stored pairs. Additional studies use 7 or 41 human-ranked pairs, crop the loser target surface before encoding, reverse the winner/loser labels, or keep \textcolor{black}{the entire large set} while adding 10\% or 30\% of randomly ordered alternatives.
In particular, synthetic cropping changes the loser clean latent and is therefore not equivalent to reordering two noise realizations of the same GT latent.

\noindent\textbf{Optimization.}
First-stage checkpoints initialize from released PartCrafter weights, while the continued \textcolor{black}{large-set} checkpoint initializes from the preceding first-stage model. Fine-tuning updates only the PartCrafter DiT $\theta$. The image encoder, VAE encoder and decoder, and a stage-initial reference copy of the DiT remain frozen. The reported models use $\beta=0.10$ and combine the reference-relative pairwise loss with target-part and all-part reconstruction. Matched reconstruction-only controls retain their corresponding training configurations but set $\lambda_{\mathrm{pair}}=0$. The primary control matches the \textcolor{black}{medium-set} first stage, and the secondary control matches the continued \textcolor{black}{large-set} setting. We use $(\lambda_{\mathrm{pair}},\lambda_{\mathrm{part}},\lambda_{\mathrm{all}})=(0.40,0.30,0.05)$,
AdamW with learning rate $3\times10^{-4}$, batch size eight, fp16, EMA, and
classifier-free-guidance dropout $0.1$ on a single GPU.
Supplementary Sec.~S2 provides the complete settings.

\noindent\textbf{Evaluation Protocol.}
For each of the three source datasets, Objaverse \cite{deitke2023objaverse}, Amazon Berkeley Objects (ABO) \cite{collins2022abo}, and ShapeNet \cite{chang2015shapenet}, we construct one fixed evaluation list shared by all methods. Before evaluation, each list is filtered against the union of object identities used by all fine-tuning configurations. No training input image, GT mesh, or target-part annotation is therefore reused for scoring. Because PartObjaverse-Tiny is derived from Objaverse, the held-out Objaverse list is the closest to the training distribution. ABO and ShapeNet test transfer across changes in source provenance, category composition, and part statistics. Multi-part inputs are the primary evaluation because they require both component generation and part separation. Single-part inputs are included as an auxiliary case.

Following PartCrafter's evaluation protocol \cite{lin2025partcrafter}, we report $L_2$ Chamfer Distance (CD) \cite{fan2017pointset} and F-score (FS) \cite{tatarchenko2019singleview} at the object level (Obj-CD/Obj-FS) and matched-part level (Part-CD/Part-FS). In this paper, \emph{target part} denotes the part indexed by $p$ during training, whereas \emph{matched part} denotes a predicted part after assignment to a GT part during evaluation. Across the main-paper tables, bold metric values improve on the corresponding pretrained result, and underlines mark the best fine-tuned setting for each metric.
Part metrics sample 40{,}960 surface points per mesh, build a GT-by-prediction $L_2$ CD cost
matrix, and solve the rectangular one-to-one assignment exactly; an unmatched GT part takes
penalty $\rho=1.25\max(1,\max C)$ and $\mathrm{FS}=0$.
Supplementary Sec.~S2 defines point-cloud normalization, part assignment, and the FS threshold, which is set to $0.1$ times the normalized reference bounding-box diagonal.

\subsection{Part-Aware Generation From Limited GT}
\label{sec:overall_results}
\label{sec:baseline_controls}
\noindent\textbf{Performance Across Part Counts.}
Table~\ref{tab:baseline_controls} shows that absolute F-score declines as the requested decomposition grows from one part to 5--8 parts, indicating greater compositional difficulty. Ours improves both object and matched-part F-score over pretrained PartCrafter in every group and gives the strongest matched-part result for both 2--4 and 5--8 parts. SFT gives the best one-part matched-part metrics, but this advantage disappears for multi-part inputs, where more components must be generated and separated. The 5--8-part result therefore indicates that target-part preference learning remains effective under more complex decompositions; on the 2--8-part aggregate, matched-part F-score improves by about 5\% over the pretrained model alongside object-level gains. Figure~\ref{fig:prediction_examples_compact} shows representative improvements in part completeness and separation.

\label{sec:direct_supervision}
\noindent\textbf{DPO Versus SFT.}
Both DPO training and SFT use part-level GT and update the same DiT, but organize supervision differently. SFT directly reconstructs 140 GT object samples with PartCrafter's standard flow-matching objective. The medium setting adds three sources of supervision: a reference-relative comparison between two noised trajectories, target-part GT reconstruction, and all-part GT reconstruction. The resulting matched-part F-score is about 4\% higher than the reported SFT result on the common held-out Objaverse list. This result compares the two complete training procedures. Because their data organization and objectives differ, it should not be interpreted as a one-factor comparison between DPO and SFT.

\noindent\textbf{Cross-Dataset Transfer.}
The cross-dataset results reveal a limitation of our work. Global shape improves on ABO, but the matched-part advantage does not transfer consistently to ABO or ShapeNet. Freezing the image encoder, 3D VAE, and reference model keeps the representation close to pretrained PartCrafter, whereas the pairwise update targets block $p$ using PartObjaverse-Tiny geometry and stored trajectory order. This source-domain supervision may preserve coarse object structure more readily than target-block behavior under distribution shift. The similar behavior of direct SFT suggests that this transfer limitation is not specific to the pairwise term.
Supplementary Sec.~S3 (Table~S1) reports the complete ABO and ShapeNet pair-construction results, while Sec.~S4 (Table~S2) reports cross-dataset results by input part count.

\subsection{Effect Of Training-Set Scale}
\label{sec:pair_scale}
\noindent\textbf{Comparison Across Scales.}
Family B in Table~\ref{tab:construction_controls} compares the small, medium, and large settings. Their results are non-monotonic. The small setting does not improve matched-part quality over the pretrained reference. The medium and large settings both improve over the pretrained reference, with the medium setting slightly ahead on matched-part metrics. Training-set size alone therefore does not explain the observed differences; geometric and category coverage may also contribute.

The cross-dataset results expose a related coverage trade-off. The \textcolor{black}{medium set} transfers local part quality more effectively to ShapeNet, whereas the \textcolor{black}{large set} gives better global-shape metrics on ABO. Because the \textcolor{black}{large-set} model is a continuation checkpoint with refreshed seeds and the \textcolor{black}{small and medium} models are first-stage checkpoints, these results show a non-monotonic trend among the reported checkpoints rather than a universal optimum at the \textcolor{black}{medium scale}.

\noindent\textbf{Non-Target Parts.}
Because the seed pair perturbs every part block, improving the target part could cost the others. Supplementary Sec.~S6 tests this on 59 held-out multi-part ShapeNet objects: the medium setting raises the pretrained weakest-part F-score from 0.182 to 0.427 and the all-parts mean by 0.03, while its remaining-parts drop of 0.046 is not significant (bootstrap 95\% CI $[-0.110,0.018]$); only the small setting degrades the remaining parts markedly.

\subsection{Alternative Pair Construction}
\label{sec:pair_source}
\noindent\textbf{Human-Ranked Pairs.}
Within family C, C1 and C2 replace metric ordering with visual comparison between generated candidates. The small, screened C1 set improves upon A1, whereas the larger C2 set degrades sharply. The objective consumes only the comparison direction, so ambiguous differences or inconsistent target-part judgments can weaken the signal. C1, which combines part-level metric checks with multiview inspection, suggests that comparison clarity can matter more than count.

\noindent\textbf{Synthetic-Crop Pairs.}
Settings C3 and C4 alter loser construction rather than the ordering source. Synthetic cropping removes target-part surface points and re-encodes the loser part set, changing its clean latent. In this setup, neither cropped-target control improves on the pretrained baseline or the metric-ranked small setting. Increasing the crop severity also provides no consistent benefit.

\begin{table}[!t]
\centering
\caption{Pair-construction, order-perturbation, and augmentation results on held-out Objaverse. The table reports 2--8-part averages for families A (pretrained reference), B (training-set scale), C (alternative construction), and D (pair perturbation).
Supplementary Table~S1 gives complete cross-dataset results.}
\label{tab:construction_controls}
\resulttablesize
\resulttablespacing{1.5pt}
\begin{tabular*}{\columnwidth}{@{\extracolsep{\fill}}llrrrr@{}}
\toprule
Family & Setting & Obj-CD $\downarrow$ & Obj-FS $\uparrow$ & Part-CD $\downarrow$ & Part-FS $\uparrow$ \\
\midrule
\textbf{A} & \textbf{A1} Pretrained & 0.2724 & 0.5680 & 0.2875 & 0.5453 \\
\midrule
\multirow{3}{*}{\textbf{B}} & \textbf{B1} Small & \textbf{0.2626} & 0.5328 & 0.3157 & 0.4545 \\
 & \textbf{B2} Medium & \textbf{\underline{0.1986}} & \textbf{\underline{0.6958}} & \textbf{\underline{0.2542}} & \textbf{\underline{0.5725}} \\
 & \textbf{B3} Large & \textbf{0.2625} & \textbf{0.5881} & \textbf{0.2716} & \textbf{0.5707} \\
\midrule
\multirow{4}{*}{\textbf{C}} & \textbf{C1} Human 7 & \textbf{0.2603} & \textbf{0.5854} & \textbf{0.2805} & \textbf{0.5573} \\
& \textbf{C2} Human 41 & 0.5127 & 0.1745 & 0.5616 & 0.1285 \\
 & \textbf{C3} Crop 30\% & 0.3529 & 0.4127 & 0.3821 & 0.3398 \\
 & \textbf{C4} Crop 50\% & 0.3636 & 0.3721 & 0.4222 & 0.2836 \\
\midrule
\multirow{5}{*}{\textbf{D}} & \textbf{D1} Reversed 10\% & 0.3848 & 0.3238 & 0.4336 & 0.2690 \\
 & \textbf{D2} Reversed 30\% & 0.3628 & 0.3709 & 0.4267 & 0.2858 \\
 & \textbf{D3} Reversed 50\% & 0.3738 & 0.3487 & 0.4480 & 0.2579 \\
 & \textbf{D4} Random add. (+80) & \textbf{0.2461} & 0.5330 & 0.3122 & 0.4319 \\
 & \textbf{D5} Random add. (+239) & 0.3542 & 0.3554 & 0.4398 & 0.2364 \\
\bottomrule
\end{tabular*}
\end{table}

\subsection{Pair Perturbations}
\label{sec:ablation_reliability}
\noindent\textbf{Pair Reversal.}
Family D perturbs or augments the stored metric-ranked pairs. D1--D3 swap winner and loser labels for increasing fractions of the primary pairs. With the reference model fixed, each swap reverses the target-part comparison. All three levels reduce quality, so label flipping is not a useful augmentation for this objective. The decline also shows that stored pair order materially affects optimization.

\noindent\textbf{Random-Pair Augmentation.}
Settings D4 and D5 keep all 796 rows and append rows with randomly drawn seed pairs, enlarging the set to 876 and 1035 examples.
Unlike reversal, this changes trajectories without deliberately flipping an existing comparison. D4 gives the best object CD among family D, but its object F-score and matched-part metrics remain below pretrained PartCrafter. The benefit disappears at the larger augmentation. These controls provide no evidence that random-pair augmentation adds useful supervision.

\subsection{Ablation Studies}
\label{sec:objective_controls}
\noindent\textbf{Effect Of $\beta$.}
The parameter $\beta$ scales the reference-relative trajectory gap in Eq.~\ref{eq:pair} and does not weight the reconstruction losses. Table~\ref{tab:beta_ablation} reports the ShapeNet $\beta$ ablation (family E), evaluated on one fixed list. Because this checkpoint series and evaluation set differ from the Objaverse loss controls, absolute values should not be compared across the two tables.

\begin{table}[!t]
\centering
\caption{Effect of $\beta$ on ShapeNet multi-part generation. Family E compares pretrained PartCrafter (E1) with fine-tuned settings E2--E5 on the same fixed list.}
\label{tab:beta_ablation}
\resulttablespacing{3pt}
{\resulttablesize
\begin{tabular*}{0.98\columnwidth}{@{\extracolsep{\fill}}lrrrr@{}}
\toprule
Setting & Obj-CD $\downarrow$ & Obj-FS $\uparrow$ & Part-CD $\downarrow$ & Part-FS $\uparrow$ \\
\midrule
\textbf{E1} Pretrained & 0.1814 & 0.7130 & 0.3262 & 0.4457 \\
\midrule
\textbf{E2} $\beta=0.05$ & 0.3003 & 0.4293 & 0.3872 & 0.2931 \\
\textbf{E3} $\beta=0.10$ & \textbf{\underline{0.1776}} & \textbf{\underline{0.7237}} & 0.3288 & \textbf{0.4513} \\
\textbf{E4} $\beta=0.20$ & 0.2249 & 0.6194 & \textbf{\underline{0.3049}} & \textbf{\underline{0.4564}} \\
\textbf{E5} $\beta=0.40$ & 0.2755 & 0.5033 & 0.3580 & 0.3368 \\
\bottomrule
\end{tabular*}
}
\end{table}

The result is non-monotonic. E2 falls below E1 on all four metrics. E3 gives the best object metrics and improves Part-FS, whereas E4 gives the best matched-part metrics at the cost of object quality. E5 degrades both metric families. The reported configurations therefore set $\beta=0.10$, the most balanced value in this ablation rather than a universal optimum.

\noindent\textbf{Pairwise-Term Control.}
Table~\ref{tab:objective_controls} reports family F on the held-out Objaverse set. F2/F3 and F4/F5 are matched pairs. Setting $\lambda_{\mathrm{pair}}=0$ retains winner-trajectory target-part and all-part reconstruction, along with the corresponding data, initialization stage, optimizer, budget, and evaluation, but removes the reference-relative comparison. 

\begin{table}[!t]
\centering
\caption{Effect of the pairwise term on held-out Objaverse. Family F evaluates 2--8-part inputs with pretrained F1 as reference. F2/F3 and F4/F5 respectively share the medium- and large-scale configurations. ``Recon.'' sets $\lambda_{\mathrm{pair}}=0$, while ``full'' includes the pairwise term. F3 is evaluated in a separate run from B2 in Table~\ref{tab:construction_controls}.}
\label{tab:objective_controls}
\resulttablespacing{2pt}
{\resulttablesize
\begin{tabular*}{0.98\columnwidth}{@{\extracolsep{\fill}}lrrrr@{}}
\toprule
Setting & Obj-CD $\downarrow$ & Obj-FS $\uparrow$ & Part-CD $\downarrow$ & Part-FS $\uparrow$ \\
\midrule
\textbf{F1} Pretrained & 0.2724 & 0.5680 & 0.2875 & 0.5453 \\
\midrule
\textbf{F2} \textcolor{black}{Medium}, recon. & \textbf{0.2650} & 0.5179 & 0.3001 & 0.4552 \\
\textbf{F3} \textcolor{black}{Medium}, full & \textbf{\underline{0.1941}} & \textbf{\underline{0.7036}} & \textbf{\underline{0.2624}} & \textbf{0.5488} \\
\midrule
\textbf{F4} \textcolor{black}{Large}, recon. & 0.2819 & 0.4878 & 0.3328 & 0.4061 \\
\textbf{F5} \textcolor{black}{Large}, full & \textbf{0.2625} & \textbf{0.5881} & \textbf{0.2716} & \textbf{\underline{0.5707}} \\
\bottomrule
\end{tabular*}
}
\end{table}

Adding the pairwise term improves every metric in both matched comparisons. In the primary \textcolor{black}{medium-set} control, it reduces object CD by about 27\% relative to reconstruction-only training. Paired bootstrap intervals over the 200 shared inputs also exclude zero in the favorable direction for every metric (Supplementary Sec.~S5). The continued \textcolor{black}{large-set} control shows the same directional pattern. These matched controls isolate the contribution of the pairwise term within the reported objective.

\FloatBarrier

\section{Limitations And Future Work}
\label{sec:limitations}
\textcolor{black}{Our gains are strongest on Objaverse and transfer less consistently to ABO and ShapeNet, whose category and decomposition statistics differ from PartObjaverse-Tiny. Broader pair collection across categories, part counts, thin structures, and target-domain samples may improve robustness. The method still requires part-level GT geometry as the reconstruction target, and offline pair ordering makes performance sensitive to candidate coverage, ordering reliability, and the selection criterion. Closing the loop by regenerating and rescoring candidates from the adapted model, while retaining a smaller GT subset for reconstruction, is a natural next step. Finally, because the preference loss acts on block $p$ but updates a DiT shared by all blocks, stronger local alignment can shift whole-object geometry. The $\beta$ sweep shows that the best local and global settings differ. All-part reconstruction constrains but does not eliminate this coupling; part-adaptive weights or lightweight adapters could address it.}

\FloatBarrier

\section{Conclusion}
\textcolor{black}{We adapt a pretrained part-aware generator with metric-ranked target-part seed pairs and a reference-relative target-block preference loss. Target-part and all-part GT flow-matching losses provide absolute supervision, while only the DiT is updated. Among the tested training-set scales, the medium-scale setting gives the strongest held-out Objaverse result, raising matched-part F-score by approximately 5\% over pretrained PartCrafter and exceeding the reported direct-SFT baseline on the multi-part aggregate. Matched controls at the medium and large scales show that adding the pairwise term improves upon reconstruction-only training. Results across the three scales are non-monotonic, while the remaining controls show sensitivity to pair ordering, target-part coverage, and objective balance. Because gains transfer less consistently to ABO and ShapeNet, we do not claim broad cross-dataset improvement.}


\FloatBarrier

{\footnotesize
\bibliographystyle{ieee}
\bibliography{references_corrected}
}

\end{document}


\dictafinalcopy
\maketitle
\thispagestyle{empty}
\pagestyle{empty}
\showsuppauthorstrue
\begin{multicols}{2}

\section{Additional Related Work}
This section expands the related-work context omitted from the main paper. It covers \emph{image-to-3D and text-to-3D generation}, \emph{part-aware 3D generation}, \emph{editable and articulated assets}, and \emph{preference and reward alignment}.

\noindent\textbf{Image-to-3D And Text-to-3D Generation.}
Recent 3D generation methods have made substantial progress in synthesizing object-level geometry from text, images, or sparse views. Score-distillation methods optimize 3D representations with guidance from pretrained two-dimensional (2D) diffusion models \cite{poole2022dreamfusion,lin2023magic3d}, while sparse-view, feed-forward reconstruction, and large-scale shape-generation pipelines improve single-image or few-view reconstruction efficiency \cite{liu2023zero1to3,hong2023lrm,li2025triposg}. These methods provide strong object-level generation baselines, but their objectives do not by themselves guarantee that local semantic components are separated, geometrically complete, or directly editable.

\noindent\textbf{Part-Aware 3D Generation.}
Part-aware generation methods can be organized into several broad families. Decompose-then-reconstruct pipelines can segment multi-view renderings, complete occluded part views, and reconstruct the resulting parts \cite{chen2025partgen}. Native part-generation methods instead represent parts explicitly as latent or token groups, using compositional part latents, contextual part latents, or autoregressive part discovery \cite{lin2025partcrafter,dong2025from,chen2025autopartgen}. Other work emphasizes semantic decoupling and structural cohesion \cite{yang2025omnipart}, efficient multi-part packing \cite{tang2025efficient}, or retrieval-augmented part editing \cite{li2026partrag}. These works primarily improve the generator architecture, representation, or conditioning interface. We instead keep the pretrained PartCrafter backbone as the starting point and study an additional part-level training signal.

\noindent\textbf{Editable And Articulated Assets.}
A related line studies whether generated parts are not only visible but also functional, editable, or usable in downstream asset pipelines. Articulated-generation methods infer part graphs, joints, axes, or motion ranges \cite{liu2024singapo,liu2026pact}.
Missing, fused, or geometrically weak parts provide unreliable structure for later editing or articulation.

\noindent\textbf{Preference And Reward Alignment.}
Direct Preference Optimization (DPO) compares ordered samples relative to a frozen reference \cite{rafailov2023direct}, and diffusion adaptations use denoising-space surrogates when output likelihoods are not tractable \cite{wallace2024diffusiondpo}. Noisy-feedback work motivates controlled order perturbations \cite{raychowdhury2024robustdpo}. In 3D vision, Circular-DPO transfers preferences across multi-stage 3D generation \cite{li2026circulardpo}, DreamDPO ranks paired renderings with reward-model or multimodal-model feedback \cite{zhou2025dreamdpoaligningtextto3dgeneration}, and Mesh-RFT applies local face-level reinforcement feedback \cite{liu2025meshrftenhancingmeshgeneration}.
Our target-part loss evaluates one indexed PartCrafter Diffusion Transformer (DiT) latent block rather than a semantic class label supplied to the generator. Each primary metric-ranked training example stores a preference pair selected before fine-tuning. The two seeds create different Gaussian noise for the same ground-truth (GT) compositional latent. Candidate-quality information determines the offline pair order but is not optimized as an online reward. Human-ranked controls replace automatic pair annotation with visual inspection of generated outputs.

\section{Experimental Details}
For reproducibility and interpretation, we first define the automatic target-part quality score and then report the training data, controls, configurations, evaluation implementation, and comparability details.

\begin{table*}[!t]
\centering
\begin{minipage}{0.98\textwidth}
\noindent\textbf{Notation.}
The table follows Eqs.~1--5 in the main paper. For $i\in\{w,\ell\}$, $w$ and $\ell$ index the winner and loser (preferred and rejected) trajectories of one stored pair, respectively. A subscript $q$ selects a generic part block, whereas $p$ selects the target part for that example.
\end{minipage}
\vspace{3pt}

\footnotesize
\setlength{\tabcolsep}{4pt}
\renewcommand{\arraystretch}{1.05}
\begin{tabular}{@{}p{0.21\textwidth}p{0.75\textwidth}@{}}
\toprule
Symbol & Definition \\
\midrule
\multicolumn{2}{@{}l}{\textit{Inputs And Pair Indexing}} \\
\addlinespace[1pt]
$x$ & Input image. \\
$\mathcal G=\{g_q\}_{q=1}^{P}$ & Indexed set of $P$ ground-truth (GT) part surfaces. \\
$P,\ q,\ p$ & Requested part count, generic part index, and target-part index for one training example. \\
$w,\ \ell,\ i$ & Winner trajectory, loser trajectory, and selector $i\in\{w,\ell\}$. \\
$s^i$ & Stored random seed (superscripted $s$) that instantiates the noise tensor for trajectory $i$. \\
\addlinespace[2pt]
\multicolumn{2}{@{}l}{\textit{Encoders And Latent Variables}} \\
\addlinespace[1pt]
$E_{\mathrm{img}},\ h_x,\ c$ & Frozen image encoder, visual tokens $h_x=E_{\mathrm{img}}(x)$, and image condition supplied to the DiT. \\
$q_\phi$ & Frozen variational autoencoder (VAE) posterior used to sample the part latents. \\
$N,\ d$ & Tokens per part and token feature width. Each compositional latent has shape $P\times N\times d$. \\
$z_0^i,\ z_{0,q}^i,\ z_{0,p}^i$ & Clean compositional latent for trajectory $i$, its generic part-$q$ block, and its target-part-$p$ block. The unsuperscripted $z_0$ is the single-trajectory form. In the primary construction, $z_0^w=z_0^\ell=z_0$. \\
$\epsilon^i,\ z_t^i,\ u^i$ & Seed-determined Gaussian noise, noised latent, and reverse-velocity target for trajectory $i$. Each has shape $P\times N\times d$. \\
\addlinespace[2pt]
\multicolumn{2}{@{}l}{\textit{Flow Matching And Preference Objective}} \\
\addlinespace[1pt]
$t,\ \sigma_t$ & Sampled timestep and scheduler noise level in the flow path. \\
$\xi^i$ & Trajectory tuple $(z_0^i,\epsilon^i,c)$. \\
$\hat u_{\theta,q}(z_t^i,t;c)$ & DiT-predicted reverse-velocity block for part $q$ of trajectory $i$. \\
$\operatorname{MSE}_t(\xi^i;\theta),\ \mathcal L_{\mathrm{CFM}}$ & Generic flow-matching mean-squared error (MSE) over the $P\times N\times d$ compositional latent and its conditional-flow-matching expectation. \\
$\operatorname{MSE}_{t}^{q}(\xi^i;\theta)$ & Block-$q$ MSE averaged over its $N$ tokens and $d$ features. Setting $q=p$ gives the target-part error. \\
$\theta,\ \theta_{\mathrm{ref}}$ & Trainable DiT and frozen DiT copied at the start of the current training stage. \\
$r_i$ & Reference-relative target-block error reduction, $\operatorname{MSE}_{t}^{p}(\xi^i;\theta_{\mathrm{ref}})-\operatorname{MSE}_{t}^{p}(\xi^i;\theta)$. \\
$\sigma(\cdot),\ \beta$ & Logistic sigmoid in the pairwise logit (distinct from $\sigma_t$) and its scale. \\
$\ell_{\mathrm{pair}},\ \mathcal L_{\mathrm{pair}}$ & Per-example, per-timestep target-block preference loss and its training-distribution expectation. \\
$\mathcal L_{\mathrm{target}},\ \mathcal L_{\mathrm{all}}$ & Winner-trajectory target-part reconstruction mean and all-part reconstruction averaged over all winner-trajectory part blocks. \\
$s,\ S_{\mathrm{warm}},\ \mathcal L_s$ & Training-update index, reconstruction-only warm-up duration, and loss at update $s$. \\
$\lambda_{\mathrm{pair}},\lambda_{\mathrm{part}},\lambda_{\mathrm{all}}$ & Weights of the pairwise, target-part reconstruction, and all-part reconstruction terms after warm-up. \\
\addlinespace[2pt]
\multicolumn{2}{@{}l}{\textit{Offline Ranking And Evaluation}} \\
\addlinespace[1pt]
$k,\ y_k$ & Candidate-seed index and its generated output. \\
$S_{\mathrm{auto,part}},\ w_\cdot$ & Automatic target-part ranking score and its nonnegative metric weights. \\
$m_j,\ C_{qj},\ \pi^\star$ & Valid predicted part, GT-to-prediction cost, and optimal one-to-one assignment. \\
$\rho,\ \gamma_{\mathrm{unmatch}}$ & Unmatched-part penalty and its multiplier. \\
\bottomrule
\end{tabular}
\end{table*}

\noindent\textbf{Automatic Target-Part Quality Score.}
For candidate output $y_k$ and indexed target part $p$, the F-score (FS), Chamfer Distance (CD), normal consistency (NC), and component quality (CQ) terms are normalized to $[0,1]$ and oriented so that larger values are better. The automatic score is
\begin{align}
S_{\mathrm{auto,part}}(y_k)={}&
w_{\mathrm{FS}}^{\mathrm{obj}}\widetilde{\mathrm{FS}}_{\mathrm{obj}}(y_k) \nonumber\\
&+w_{\mathrm{CD}}^{\mathrm{obj}}\bigl(1-\widetilde{\mathrm{CD}}_{\mathrm{obj}}(y_k)\bigr) \nonumber\\
&+w_{\mathrm{FS}}^{\mathrm{part}}\widetilde{\mathrm{FS}}_{\mathrm{part}}(y_k) \nonumber\\
&+w_{\mathrm{CD}}^{\mathrm{part}}\bigl(1-\widetilde{\mathrm{CD}}_{\mathrm{part}}(y_k)\bigr) \nonumber\\
&+w_{\mathrm{NC}}\widetilde{\mathrm{NC}}(y_k)
+w_{\mathrm{CQ}}\mathrm{CQ}(y_k),
\label{eq:supp_quality_score}
\end{align}
where the nonnegative weights sum to one. The object terms measure the complete output, while the part terms measure indexed target part $p$. Let $\mathcal V_k$ be the valid predicted parts and $\operatorname{cc}(y_{k,j})$ the number of connected components in part $j$. Component quality is the inverse mean component count,
\begin{equation}
\mathrm{CQ}(y_k)=\left(\frac{1}{|\mathcal V_k|}\sum_{j\in\mathcal V_k}\operatorname{cc}(y_{k,j})\right)^{-1}.
\label{eq:supp_component_quality}
\end{equation}
We set $\mathrm{CQ}(y_k)=0$ when $\mathcal V_k$ is empty.
The reported ranking runs assign weights $0.35/0.20$ to object FS/CD, $0.20/0.10$ to part FS/CD, $0.10$ to normal consistency, and $0.05$ to component quality. Boundary-edge ratio and part intersection-over-union are retained as scoring diagnostics but are not included in Eq.~\ref{eq:supp_quality_score}, because their ranking power was not stable on the calibration subset. Pair construction uses $S_{\mathrm{auto,part}}$ only to rank and filter candidates. The training objective consumes the stored winner/loser order, not the scalar score itself.

\noindent\textbf{Metric-Ranked Training Set.}
The source subset contains exactly 140 PartObjaverse-Tiny objects \cite{yang2024sampart3d} and 796 target-part pair examples. Objects have two to eight usable parts, and parts below a face fraction of 0.005 are removed. Each example stores the source identity, requested part count, target-part index, indexed GT geometry, and one offline-selected seed pair. The stored record supplies the pair order rather than an online scalar reward. For every primary pair, the two seeds instantiate different Gaussian noise while the clean GT compositional latent remains the same.

\noindent\textbf{Training-Set-Size Subsets.}
The 56-pair and 363-pair subsets contain 7 and 50 source objects drawn from the 140-object set. A visibility criterion selects source objects by usable-part count and minimum part fraction while maintaining category coverage. The 796-pair result is a continuation-stage checkpoint over all 140 objects with refreshed stored pairs. Each training example contains one winner/loser seed pair used to instantiate training noise. The same winner seed may occur in multiple examples paired with different valid loser seeds.

\noindent\textbf{Human And Synthetic Controls.}
The first human audit produced 94 trainable selections. After deduplicating repeated combinations of dataset, source identity, and seed pair, 41 preference pairs remain. They comprise 31 ShapeNet, 7 Amazon Berkeley Objects (ABO), and 3 Objaverse pairs. A separate pilot retained seven visually selected ShapeNet seed pairs from five source objects after metric and multiview inspection. In both studies, a human compared the generated candidate meshes and labeled their seeds as winner or loser. During fine-tuning, the inspected meshes are not encoded as training targets. The labeled seeds instantiate two noise tensors for the same indexed GT latent. Synthetic-crop controls remove exactly 30\% or 50\% of the loser target-part surface points and then independently re-encode the complete loser part set. Thus, only the target surface geometry is modified, but stochastic VAE samples for unchanged non-target surfaces can also differ from the winner trajectory. Label-reversal controls swap the winner/loser labels for exactly 10\%, 30\%, or 50\% of pairs. Random-pair augmentation controls keep all 796 metric-ranked examples and append 80 or 239 additional examples whose seed pairs are drawn and ordered at random, enlarging the training set to 876 or 1{,}035 examples (approximately 10\% or 30\% of the primary set).

\noindent\textbf{Training Configurations.}
The 56- and 363-pair first-stage checkpoints initialize from released PartCrafter weights. The 796-pair checkpoint continues from its first-stage model, and its frozen reference DiT is copied from that stage-initial checkpoint. All three scale settings use AdamW \cite{loshchilov2019decoupled} with $(\beta_1,\beta_2)=(0.9,0.999)$, weight decay $0.01$, $\epsilon=10^{-8}$, learning rate $3\times10^{-4}$, and a constant schedule after a brief warmup. Training uses 16-bit floating point (fp16), gradient accumulation of one, gradient checkpointing, an exponential moving average (EMA), batch size eight per graphics processing unit (GPU), classifier-free guidance dropout $0.1$, and the reverse-velocity objective. We set $\beta=0.10$ and $(\lambda_{\mathrm{pair}},\lambda_{\mathrm{part}},\lambda_{\mathrm{all}})=(0.40,0.30,0.05)$. Each primary stored pair has unit example weight, pair margins are disabled, and a short reconstruction-only warmup precedes the pairwise term. The image encoder and complete 3D VAE remain frozen, and only the DiT is optimized. For each training example, we sample one timestep from a logit-normal distribution with mean 0 and standard deviation 1, map it to the 1,000-step scheduler, and apply the same timestep to all $P$ part blocks \cite{esser2024scaling}. The primary reconstruction-only control matches the 363-pair first-stage setting and its 300-update budget while setting $\lambda_{\mathrm{pair}}=0$. A secondary control analogously matches the continued 796-pair setting. Within each comparison, the full and reconstruction-only checkpoints use the same training examples, initialization stage, reconstruction terms, optimizer configuration, training budget, and fixed evaluation configuration. Direct SFT instead initializes from released weights, uses 140 GT object samples, AdamW with learning rate $1\times10^{-4}$ and batch size eight, and optimizes PartCrafter's standard flow-matching objective without an ordered pair or pairwise term. The human, crop, order-reversal, and random-pair augmentation controls use their checkpoint-specific data constructions with learning rate $3\times10^{-4}$ and batch size eight. Training uses one GPU.

\noindent\textbf{Evaluation Implementation.}
For each data source, all methods share one fixed evaluation list constructed after removing the union of object identities appearing in any fine-tuning configuration. This separation is applied before rendering and metric aggregation, so the evaluation inputs, GT meshes, and target-part annotations are not shared with the training examples. The lists are drawn from Objaverse \cite{deitke2023objaverse}, ABO \cite{collins2022abo}, and ShapeNet \cite{chang2015shapenet}. For object metrics, predicted and reference point clouds are independently centered and scaled by their maximum spatial extent before $L_2$ Chamfer Distance (CD) \cite{fan2017pointset} and F-score (FS) \cite{tatarchenko2019singleview} are computed, following the metrics used by PartCrafter \cite{lin2025partcrafter}. For part metrics, the implementation samples 40,960 surface points per mesh, forms a GT-by-valid-prediction cost matrix from $L_2$ CD, and solves the rectangular one-to-one assignment exactly. A GT part may remain unmatched with penalty
$\rho=\gamma_{\mathrm{unmatch}}\max(1,\max C)$, where
$\gamma_{\mathrm{unmatch}}=1.25$ in all evaluations. Each matched pair is then independently normalized and scored, and the part scores are averaged for that object. FS uses a threshold equal to $0.1$ times the normalized reference bounding-box diagonal. The aggregation excludes failed evaluations and samples assigned part CD of at least 100. The one-part lists contain 200 inputs per dataset. The multi-part lists contain 200 Objaverse inputs (135 with 2--4 parts and 65 with 5--8), 52 ABO inputs (all with 2--4 parts), and 200 ShapeNet inputs (108 with 2--4 parts and 92 with 5--8).

\noindent\textbf{Part-Matching Pseudocode.}
\begin{quote}
\small
\textbf{Input} Predicted meshes $\{m_j\}_{j=1}^{M}$, GT part surfaces
$\{g_q\}_{q=1}^{Q}$, and unmatched-cost multiplier
$\gamma_{\mathrm{unmatch}}>1$.\\
\textbf{Output} Mean matched-part Chamfer Distance (CD) and F-score (FS).\\[2pt]

1. Remove invalid predicted meshes and uniformly sample a point cloud from each
remaining mesh and GT part surface.\\

2. Construct the assignment cost matrix
\[
C_{qj}=\operatorname{CD}_{L_2}(g_q,m_j).
\]

3. Set the unmatched-part penalty to
\[
\rho
=
\gamma_{\mathrm{unmatch}}
\max\!\left(1,\max_{q,j}C_{qj}\right),
\]
and solve the partial one-to-one assignment
\[
\pi^\star
=
\arg\min_{\pi}
\left[
\sum_{(q,j)\in\pi} C_{qj}
+
\rho\bigl(Q-|\pi|\bigr)
\right],
\]
where each GT or predicted part appears in at most one matched pair.\\

4. For each $(q,j)\in\pi^\star$, normalize the two point clouds according to
the part-level evaluation protocol and compute their CD and FS. For every
unmatched GT part, set $\mathrm{CD}_q=\rho$ and $\mathrm{FS}_q=0$.\\

5. Return
\[
\begin{aligned}
\mathrm{CD}_{\mathrm{part}}
&=
\frac{1}{Q}\sum_{q=1}^{Q}\mathrm{CD}_q,
\\
\mathrm{FS}_{\mathrm{part}}
&=
\frac{1}{Q}\sum_{q=1}^{Q}\mathrm{FS}_q.
\end{aligned}
\]
Extra predicted parts are not included in these GT-referenced averages.
\end{quote}

\noindent\textbf{Comparability.}
The scale, human-inspected, synthetic-crop, and order-perturbation studies use the checkpoint-specific settings reported above. The reconstruction-only controls in Sec.~\ref{sec:supp_objective_control} are matched within the 363- and 796-pair settings and differ from their corresponding full objectives by disabling $\mathcal L_{\mathrm{pair}}$. The 363-pair comparison is the primary attribution test because it corresponds to the strongest main setting. The 796-pair comparison provides a continued-stage check. These controls isolate the contribution of the pairwise term more directly than comparisons across training-set constructions.

\end{multicols}
\section{Full Pair-Construction Results}
Table~\ref{tab:supp_main_results} reports the data-construction and order-perturbation results omitted from the compact main table. It separates the number of training pairs, the source of the ordering, rejected-target construction, label reversal, and random-pair augmentation.
\begin{table}[!ht]
\centering
\caption{Cross-dataset comparison of pair-construction and order-perturbation controls. Metric-ranked and human-ranked pairs use the same GT clean latent for both seeds. Synthetic cropping changes the loser clean latent, label reversal swaps the winner/loser labels, and random-pair augmentation appends randomly ordered seed pairs to the fixed 796-example set, so the GT latents are unchanged while the number of training examples grows.}
\label{tab:supp_main_results}
\resulttablesize
\resulttablespacing{0.5pt}
\resizebox{\textwidth}{!}{%
\begin{tabular*}{1.035\textwidth}{@{\extracolsep{\fill}}llrrrrrrrrrrrr@{}}
\toprule
Method & Setting & \multicolumn{4}{c}{\textbf{Objaverse}} & \multicolumn{4}{c}{\textbf{ABO}} & \multicolumn{4}{c}{\textbf{ShapeNet}} \\
\cmidrule(lr){3-6}\cmidrule(lr){7-10}\cmidrule(lr){11-14}
 & & Obj-CD $\downarrow$ & Obj-FS $\uparrow$ & Part-CD $\downarrow$ & Part-FS $\uparrow$ & Obj-CD $\downarrow$ & Obj-FS $\uparrow$ & Part-CD $\downarrow$ & Part-FS $\uparrow$ & Obj-CD $\downarrow$ & Obj-FS $\uparrow$ & Part-CD $\downarrow$ & Part-FS $\uparrow$ \\
\midrule
Baseline & Pretrained & 0.2724 & 0.5680 & 0.2875 & 0.5453 & 0.1520 & 0.7806 & 0.3204 & 0.4888 & 0.1814 & 0.7130 & 0.3262 & 0.4457 \\
\midrule
\multirow{3}{*}{Metric-Ranked} & B1 small & \metriccd{0.2626}{0.2724}{0.1986} & \metricf{0.5328}{0.5680}{0.6958} & \metriccd{0.3157}{0.2875}{0.2542} & \metricf{0.4545}{0.5453}{0.5725} & \metriccd{0.3000}{0.1520}{0.1382} & \metricf{0.4528}{0.7806}{0.8072} & \metriccd{0.3722}{0.3204}{0.3104} & \metricf{0.3437}{0.4888}{0.4729} & \metriccd{0.2843}{0.1814}{0.1833} & \metricf{0.4442}{0.7130}{0.7118} & \metriccd{0.3612}{0.3262}{0.2872} & \metricf{0.3500}{0.4457}{0.4688} \\
 & B2 medium & \metriccd{0.1986}{0.2724}{0.1986} & \metricf{0.6958}{0.5680}{0.6958} & \metriccd{0.2542}{0.2875}{0.2542} & \metricf{0.5725}{0.5453}{0.5725} & \metriccd{0.2233}{0.1520}{0.1382} & \metricf{0.6560}{0.7806}{0.8072} & \metriccd{0.3104}{0.3204}{0.3104} & \metricf{0.4707}{0.4888}{0.4729} & \metriccd{0.2099}{0.1814}{0.1833} & \metricf{0.6500}{0.7130}{0.7118} & \metriccd{0.2872}{0.3262}{0.2872} & \metricf{0.4688}{0.4457}{0.4688} \\
 & {B3 large} & \metriccd{0.2625}{0.2724}{0.1986} & \metricf{0.5881}{0.5680}{0.6958} & \metriccd{0.2716}{0.2875}{0.2542} & \metricf{0.5707}{0.5453}{0.5725} & \metriccd{0.1382}{0.1520}{0.1382} & \metricf{0.8072}{0.7806}{0.8072} & \metriccd{0.3215}{0.3204}{0.3104} & \metricf{0.4729}{0.4888}{0.4729} & \metriccd{0.1833}{0.1814}{0.1833} & \metricf{0.7118}{0.7130}{0.7118} & \metriccd{0.3344}{0.3262}{0.2872} & \metricf{0.4396}{0.4457}{0.4688} \\
\midrule
\multirow{2}{*}{Human-Ranked} & 41 pairs & \metriccd{0.5127}{0.2724}{0.2603} & \metricf{0.1745}{0.5680}{0.5854} & \metriccd{0.5616}{0.2875}{0.2805} & \metricf{0.1285}{0.5453}{0.5573} & \metriccd{0.4440}{0.1520}{0.1400} & \metricf{0.2215}{0.7806}{0.8021} & \metriccd{0.5293}{0.3204}{0.3096} & \metricf{0.1531}{0.4888}{0.5006} & \metriccd{0.4942}{0.1814}{0.1725} & \metricf{0.1579}{0.7130}{0.7386} & \metriccd{0.5557}{0.3262}{0.3267} & \metricf{0.1284}{0.4457}{0.4493} \\
 & 7 pairs & \metriccd{0.2603}{0.2724}{0.2603} & \metricf{0.5854}{0.5680}{0.5854} & \metriccd{0.2805}{0.2875}{0.2805} & \metricf{0.5573}{0.5453}{0.5573} & \metriccd{0.1400}{0.1520}{0.1400} & \metricf{0.8021}{0.7806}{0.8021} & \metriccd{0.3096}{0.3204}{0.3096} & \metricf{0.5006}{0.4888}{0.5006} & \metriccd{0.1725}{0.1814}{0.1725} & \metricf{0.7386}{0.7130}{0.7386} & \metriccd{0.3267}{0.3262}{0.3267} & \metricf{0.4493}{0.4457}{0.4493} \\
\midrule
\multirow{2}{*}{Synthetic Crop} & +crop 30\% & \metriccd{0.3529}{0.2724}{0.3529} & \metricf{0.4127}{0.5680}{0.4127} & \metriccd{0.3821}{0.2875}{0.3821} & \metricf{0.3398}{0.5453}{0.3398} & \metriccd{0.3319}{0.1520}{0.3319} & \metricf{0.4355}{0.7806}{0.4355} & \metriccd{0.4145}{0.3204}{0.4145} & \metricf{0.3072}{0.4888}{0.3072} & \metriccd{0.3681}{0.1814}{0.3514} & \metricf{0.3531}{0.7130}{0.3717} & \metriccd{0.4490}{0.3262}{0.4346} & \metricf{0.2541}{0.4457}{0.2631} \\
 & +crop 50\% & \metriccd{0.3636}{0.2724}{0.3529} & \metricf{0.3721}{0.5680}{0.4127} & \metriccd{0.4222}{0.2875}{0.3821} & \metricf{0.2836}{0.5453}{0.3398} & \metriccd{0.3610}{0.1520}{0.3319} & \metricf{0.3950}{0.7806}{0.4355} & \metriccd{0.4280}{0.3204}{0.4145} & \metricf{0.2945}{0.4888}{0.3072} & \metriccd{0.3514}{0.1814}{0.3514} & \metricf{0.3717}{0.7130}{0.3717} & \metriccd{0.4346}{0.3262}{0.4346} & \metricf{0.2631}{0.4457}{0.2631} \\
\midrule
\multirow{3}{*}{Label Reversal} & 10\% reversed & \metriccd{0.3848}{0.2724}{0.3628} & \metricf{0.3238}{0.5680}{0.3709} & \metriccd{0.4336}{0.2875}{0.4267} & \metricf{0.2690}{0.5453}{0.2858} & \metriccd{0.3678}{0.1520}{0.3678} & \metricf{0.3957}{0.7806}{0.3957} & \metriccd{0.4562}{0.3204}{0.4562} & \metricf{0.2672}{0.4888}{0.2672} & \metriccd{0.3722}{0.1814}{0.3722} & \metricf{0.3478}{0.7130}{0.3478} & \metriccd{0.4289}{0.3262}{0.4289} & \metricf{0.2645}{0.4457}{0.2645} \\
 & 30\% reversed & \metriccd{0.3628}{0.2724}{0.3628} & \metricf{0.3709}{0.5680}{0.3709} & \metriccd{0.4267}{0.2875}{0.4267} & \metricf{0.2858}{0.5453}{0.2858} & \metriccd{0.4026}{0.1520}{0.3678} & \metricf{0.3161}{0.7806}{0.3957} & \metriccd{0.4783}{0.3204}{0.4562} & \metricf{0.2384}{0.4888}{0.2672} & \metriccd{0.3804}{0.1814}{0.3722} & \metricf{0.3102}{0.7130}{0.3478} & \metriccd{0.4560}{0.3262}{0.4289} & \metricf{0.2434}{0.4457}{0.2645} \\
 & 50\% reversed & \metriccd{0.3738}{0.2724}{0.3628} & \metricf{0.3487}{0.5680}{0.3709} & \metriccd{0.4480}{0.2875}{0.4267} & \metricf{0.2579}{0.5453}{0.2858} & \metriccd{0.3996}{0.1520}{0.3678} & \metricf{0.3247}{0.7806}{0.3957} & \metriccd{0.4822}{0.3204}{0.4562} & \metricf{0.2192}{0.4888}{0.2672} & \metriccd{0.3833}{0.1814}{0.3722} & \metricf{0.3178}{0.7130}{0.3478} & \metriccd{0.4650}{0.3262}{0.4289} & \metricf{0.2194}{0.4457}{0.2645} \\
\midrule
\multirow{2}{*}{Random Augmentation} & +80 (10\%) & \metriccd{0.2461}{0.2724}{0.2461} & \metricf{0.5330}{0.5680}{0.5330} & \metriccd{0.3122}{0.2875}{0.3122} & \metricf{0.4319}{0.5453}{0.4319} & \metriccd{0.2267}{0.1520}{0.2267} & \metricf{0.6041}{0.7806}{0.6041} & \metriccd{0.3287}{0.3204}{0.3287} & \metricf{0.4440}{0.4888}{0.4440} & \metriccd{0.2432}{0.1814}{0.2432} & \metricf{0.5240}{0.7130}{0.5240} & \metriccd{0.3227}{0.3262}{0.3227} & \metricf{0.3975}{0.4457}{0.3975} \\
 & +239 (30\%) & \metriccd{0.3542}{0.2724}{0.2461} & \metricf{0.3554}{0.5680}{0.5330} & \metriccd{0.4398}{0.2875}{0.3122} & \metricf{0.2364}{0.5453}{0.4319} & \metriccd{0.3549}{0.1520}{0.2267} & \metricf{0.2719}{0.7806}{0.6041} & \metriccd{0.5075}{0.3204}{0.3287} & \metricf{0.1812}{0.4888}{0.4440} & \metriccd{0.3982}{0.1814}{0.2432} & \metricf{0.2512}{0.7130}{0.5240} & \metriccd{0.5219}{0.3262}{0.3227} & \metricf{0.1458}{0.4457}{0.3975} \\
\bottomrule
\end{tabular*}
}
\end{table}
The metric-ranked medium-scale setting gives the best Objaverse matched-part metrics. The 796-pair setting gives the best ABO object values in that group, showing that focused target-part supervision and broader object coverage can favor different transfer properties. The seven-pair human study improves all four Objaverse and ABO metrics and three of four ShapeNet metrics over pretrained. ShapeNet Part CD is effectively unchanged. This result suggests that carefully selected human comparisons can provide useful supervision even with only seven pairs. Among the perturbation controls, random-pair augmentation with 80 added pairs gives the best object CD on all three datasets, but its object FS and matched-part metrics stay below pretrained PartCrafter, and the larger 239-pair augmentation falls below it on every metric.

\section{Full Part-Count Results}
Table~\ref{tab:supp_part_count_results} gives the complete cross-dataset version of main-paper Table~1, including the ABO and ShapeNet values omitted from the compact Objaverse-only presentation.
\begin{table}[!ht]
\centering
\caption{Comparison by input part count across Objaverse, ABO, and ShapeNet. Ours is the continued-stage checkpoint trained with 796 metric-ranked trajectory pairs. SFT uses 140 GT object samples and the standard flow-matching objective. The 2--8 average excludes single-part inputs and aggregates valid evaluation examples.}
\label{tab:supp_part_count_results}
\resulttablesize
\resulttablespacing{0.5pt}
\begin{tabular*}{\textwidth}{@{\extracolsep{\fill}}llrrrrrrrrrrrr@{}}
\toprule
Part Count & Method & \multicolumn{4}{c}{\textbf{Objaverse}} & \multicolumn{4}{c}{\textbf{ABO}} & \multicolumn{4}{c}{\textbf{ShapeNet}} \\
\cmidrule(lr){3-6}\cmidrule(lr){7-10}\cmidrule(lr){11-14}
 & & Obj-CD $\downarrow$ & Obj-FS $\uparrow$ & Part-CD $\downarrow$ & Part-FS $\uparrow$ & Obj-CD $\downarrow$ & Obj-FS $\uparrow$ & Part-CD $\downarrow$ & Part-FS $\uparrow$ & Obj-CD $\downarrow$ & Obj-FS $\uparrow$ & Part-CD $\downarrow$ & Part-FS $\uparrow$ \\
\midrule
\multirow{3}{*}{1 part} & Pretrained & \metricbest{0.2382}{0.2184} & \metricbest{0.6366}{0.6603} & \metricbest{0.2096}{0.1890} & \metricbest{0.6949}{0.7273} & \metricbest{0.0978}{0.0978} & \metricbest{0.8998}{0.9066} & \metricbest{0.0975}{0.0975} & \metricbest{0.9007}{0.9007} & \metricbest{0.2092}{0.2092} & \metricbest{0.6626}{0.6626} & \metricbest{0.2094}{0.2094} & \metricbest{0.6632}{0.6632} \\
 & SFT & \metriccd{0.2243}{0.2382}{0.2184} & \metricf{0.6441}{0.6366}{0.6603} & \metriccd{0.1890}{0.2096}{0.1890} & \metricf{0.7273}{0.6949}{0.7273} & \metriccd{0.0991}{0.0978}{0.0978} & \metricf{0.9066}{0.8998}{0.9066} & \metriccd{0.1043}{0.0975}{0.0975} & \metricf{0.8935}{0.9007}{0.9007} & \metriccd{0.2162}{0.2092}{0.2092} & \metricf{0.6411}{0.6626}{0.6626} & \metriccd{0.2161}{0.2094}{0.2094} & \metricf{0.6401}{0.6632}{0.6632} \\
 & \textbf{Ours} & \metriccd{0.2184}{0.2382}{0.2184} & \metricf{0.6603}{0.6366}{0.6603} & \metriccd{0.2044}{0.2096}{0.1890} & \metricf{0.7020}{0.6949}{0.7273} & \metriccd{0.1007}{0.0978}{0.0978} & \metricf{0.8883}{0.8998}{0.9066} & \metriccd{0.1015}{0.0975}{0.0975} & \metricf{0.8898}{0.9007}{0.9007} & \metriccd{0.2276}{0.2092}{0.2092} & \metricf{0.6147}{0.6626}{0.6626} & \metriccd{0.2294}{0.2094}{0.2094} & \metricf{0.6000}{0.6632}{0.6632} \\
\midrule
\multirow{3}{*}{2--4 parts} & Pretrained & \metricbest{0.2631}{0.2491} & \metricbest{0.5909}{0.6121} & \metricbest{0.2574}{0.2485} & \metricbest{0.5986}{0.6237} & \metricbest{0.1520}{0.1382} & \metricbest{0.7806}{0.8087} & \metricbest{0.3204}{0.3062} & \metricbest{0.4888}{0.4889} & \metricbest{0.1828}{0.1807} & \metricbest{0.7107}{0.7179} & \metricbest{0.3115}{0.3099} & \metricbest{0.4689}{0.4748} \\
 & SFT & \metriccd{0.2568}{0.2631}{0.2491} & \metricf{0.5945}{0.5909}{0.6121} & \metriccd{0.2620}{0.2574}{0.2485} & \metricf{0.5955}{0.5986}{0.6237} & \metriccd{0.1409}{0.1520}{0.1382} & \metricf{0.8087}{0.7806}{0.8087} & \metriccd{0.3062}{0.3204}{0.3062} & \metricf{0.4889}{0.4888}{0.4889} & \metriccd{0.1807}{0.1828}{0.1807} & \metricf{0.7178}{0.7107}{0.7179} & \metriccd{0.3099}{0.3115}{0.3099} & \metricf{0.4748}{0.4689}{0.4748} \\
 & \textbf{Ours} & \metriccd{0.2491}{0.2631}{0.2491} & \metricf{0.6121}{0.5909}{0.6121} & \metriccd{0.2485}{0.2574}{0.2485} & \metricf{0.6237}{0.5986}{0.6237} & \metriccd{0.1382}{0.1520}{0.1382} & \metricf{0.8072}{0.7806}{0.8087} & \metriccd{0.3215}{0.3204}{0.3062} & \metricf{0.4729}{0.4888}{0.4889} & \metriccd{0.1816}{0.1828}{0.1807} & \metricf{0.7179}{0.7107}{0.7179} & \metriccd{0.3199}{0.3115}{0.3099} & \metricf{0.4712}{0.4689}{0.4748} \\
\midrule
\multirow{3}{*}{5--8 parts} & Pretrained & \metricbest{0.2865}{0.2813} & \metricbest{0.5332}{0.5520} & \metricbest{0.3333}{0.3062} & \metricbest{0.4646}{0.4912} & -- & -- & -- & -- & \metricbest{0.1795}{0.1677} & \metricbest{0.7160}{0.7391} & \metricbest{0.3456}{0.3456} & \metricbest{0.4149}{0.4149} \\
 & SFT & \metriccd{0.2813}{0.2865}{0.2813} & \metricf{0.5463}{0.5332}{0.5520} & \metriccd{0.3197}{0.3333}{0.3062} & \metricf{0.4846}{0.4646}{0.4912} & -- & -- & -- & -- & \metriccd{0.1677}{0.1795}{0.1677} & \metricf{0.7391}{0.7160}{0.7391} & \metriccd{0.3497}{0.3456}{0.3456} & \metricf{0.4006}{0.4149}{0.4149} \\
 & \textbf{Ours} & \metriccd{0.2824}{0.2865}{0.2813} & \metricf{0.5520}{0.5332}{0.5520} & \metriccd{0.3062}{0.3333}{0.3062} & \metricf{0.4912}{0.4646}{0.4912} & -- & -- & -- & -- & \metriccd{0.1854}{0.1795}{0.1677} & \metricf{0.7037}{0.7160}{0.7391} & \metriccd{0.3537}{0.3456}{0.3456} & \metricf{0.3975}{0.4149}{0.4149} \\
\midrule
\multirow{3}{*}{\textbf{2--8 avg.}} & Pretrained & \metricbest{0.2724}{0.2625} & \metricbest{0.5680}{0.5881} & \metricbest{0.2875}{0.2716} & \metricbest{0.5453}{0.5707} & \metricbest{0.1520}{0.1382} & \metricbest{0.7806}{0.8087} & \metricbest{0.3204}{0.3062} & \metricbest{0.4888}{0.4889} & \metricbest{0.1814}{0.1751} & \metricbest{0.7130}{0.7269} & \metricbest{0.3262}{0.3262} & \metricbest{0.4457}{0.4457} \\
 & SFT & \metriccd{0.2667}{0.2724}{0.2625} & \metricf{0.5751}{0.5680}{0.5881} & \metriccd{0.2852}{0.2875}{0.2716} & \metricf{0.5509}{0.5453}{0.5707} & \metriccd{0.1409}{0.1520}{0.1382} & \metricf{0.8087}{0.7806}{0.8087} & \metriccd{0.3062}{0.3204}{0.3062} & \metricf{0.4889}{0.4888}{0.4889} & \metriccd{0.1751}{0.1814}{0.1751} & \metricf{0.7269}{0.7130}{0.7269} & \metriccd{0.3270}{0.3262}{0.3262} & \metricf{0.4430}{0.4457}{0.4457} \\
 & \textbf{Ours} & \metriccd{0.2625}{0.2724}{0.2625} & \metricf{0.5881}{0.5680}{0.5881} & \metriccd{0.2716}{0.2875}{0.2716} & \metricf{0.5707}{0.5453}{0.5707} & \metriccd{0.1382}{0.1520}{0.1382} & \metricf{0.8072}{0.7806}{0.8087} & \metriccd{0.3215}{0.3204}{0.3062} & \metricf{0.4729}{0.4888}{0.4889} & \metriccd{0.1833}{0.1814}{0.1751} & \metricf{0.7118}{0.7130}{0.7269} & \metriccd{0.3344}{0.3262}{0.3262} & \metricf{0.4396}{0.4457}{0.4457} \\
\bottomrule
\end{tabular*}
\end{table}
\FloatBarrier



\FloatBarrier

\section{Matched Objective Control}
\label{sec:supp_objective_control}
Main-paper Table~4 reports two matched comparisons in which the reconstruction-only checkpoint sets $\lambda_{\mathrm{pair}}=0$, while the training data, initialization stage, reconstruction terms, optimizer, training budget, and evaluation configuration are shared with its full-objective counterpart. The 363-pair F2/F3 block is a separate matched rerun and should not be identified with pair-count setting B2. Its full objective improves every metric over the reconstruction-only control. A paired bootstrap over the 200 shared inputs with 20,000 replicates gives improvement-direction 95\% confidence intervals of $[0.0577,0.0846]$ for object CD reduction, $[0.1561,0.2156]$ for object FS increase, $[0.0290,0.0465]$ for part CD reduction, and $[0.0735,0.1140]$ for part FS increase. The 796-pair F4/F5 block gives the same directional result as a corroborating continued-stage control.

\FloatBarrier


\FloatBarrier

\section{Will non-target parts degrade?}
\label{sec:supp_per_part}

\begin{table}[!ht]
\centering
\caption{Evaluation of the weakest and remaining parts on 59 paired multi-part ShapeNet objects comprising 276 GT parts. For each object, we identify the GT part with the lowest pretrained F-score and report the performance of each setting on this weakest part, the mean of the remaining parts, and the mean over all parts. The pretrained model determines the weakest-part selection and serves as the baseline for measuring improvements from DPO-guided fine-tuning. Underlined values indicate improvements over the pretrained baseline, while bold values denote the best results among our fine-tuned settings.}
\label{tab:supp_per_part}
\resulttablesize
\resulttablespacing{4pt}
\begin{tabular*}{\textwidth}{@{\extracolsep{\fill}}lrrrrrr@{}}
\toprule
\multirow{2}{*}{Setting} & \multicolumn{2}{c}{Weakest part} & \multicolumn{2}{c}{Remaining parts avg.} & \multicolumn{2}{c}{All parts avg.} \\
\cmidrule(lr){2-3}\cmidrule(lr){4-5}\cmidrule(lr){6-7}
 & CD $\downarrow$ & FS $\uparrow$ & CD $\downarrow$ & FS $\uparrow$ & CD $\downarrow$ & FS $\uparrow$ \\
\midrule
Pretrained & 0.4476 & 0.1820 & {0.2757} & {0.5299} & {0.3179} & {0.4397} \\
\midrule
B1 small   & \underline{0.3735} & \underline{0.3238} & 0.3684 & 0.3359 & 0.3712 & 0.3304 \\
B2 medium  & \textbf{\underline{0.3007}} & \textbf{\underline{0.4272}} & \underline{0.2788} & \underline{0.4841} & \textbf{\underline{0.2848}} & \textbf{\underline{0.4681}} \\
B3 large   & \underline{0.3950} & \underline{0.3039} & 0.3066 & 0.4773 & 0.3303 & 0.4286 \\
\bottomrule
\end{tabular*}
\end{table}

Table~\ref{tab:supp_per_part} shows that B2 achieves the best weakest-part performance and the
best all-parts average, improving the all-parts F-score by $0.03$. Thus, improving the weakest part does not reduce the overall
per-part performance.

However, the remaining-part F-score for our best setting B2 is $0.046$ lower than that of pretrained PartCrafter.
However, this difference is not statistically significant, with a $95\%$ bootstrap interval of
$[-0.110,+0.018]$ and $p=0.17$. Therefore, the results provide no strong evidence of systematic
degradation in non-target parts. B1 shows a larger remaining-part drop of $0.194$, but B1, B2,
and B3 differ primarily in the number of preference pairs ($56$, $363$, and $796$), suggesting
that the degradation in B1 is more likely related to insufficient training-data coverage than to
an inherent conflict between target-part improvement and the preservation of other parts.

\FloatBarrier

\begin{multicols}{2}
{\fontsize{6.5}{7.1}\selectfont
\bibliographystyle{ieee}
\bibliography{references}
}
\end{multicols}